\documentclass[10pt, conference, compsocconf]{IEEEtran}
\ifCLASSINFOpdf
\else
\fi
\usepackage[colorinlistoftodos]{todonotes}
\usepackage{algorithmic}
\usepackage[ruled,vlined,linesnumbered]{algorithm2e}
\usepackage{amsmath}
\usepackage{amssymb}
\usepackage{booktabs} 
\usepackage{xspace}

\DeclareMathOperator*{\argmax}{arg\,max}

\usepackage{xfrac}    
\usepackage{nicefrac} 
\usepackage{multirow}
\usepackage{times}
\usepackage{graphicx}
\usepackage{subcaption}
\usepackage{todonotes}
\usepackage{wrapfig}
\usepackage{hyperref}
\usepackage{url}
\usepackage{color}
\usepackage{balance}

\begin{document}
%
\title{BiasedWalk: Biased Sampling for Representation Learning on Graphs}


\author{\IEEEauthorblockN{Duong Nguyen}
\IEEEauthorblockA{Department of Computer Science and Engineering\\
UC San Diego\\
La Jolla, CA, USA\\
Email: duong@eng.ucsd.edu}
\and
\IEEEauthorblockN{Fragkiskos D. Malliaros}
\IEEEauthorblockA{Center for Visual Computing\\
CentraleSup\'{e}lec and Inria Saclay\\
Gif-Sur-Yvette, France\\
Email: fragkiskos.malliaros@centralesupelec.fr}
}


%


\maketitle

\begin{abstract}
Network embedding algorithms are able to learn latent feature representations of nodes, transforming networks into lower dimensional vector representations.
Typical key applications, which have effectively been addressed using network embeddings, include link prediction, multilabel classification and community detection. In this paper, we propose BiasedWalk, a scalable, unsupervised feature learning algorithm that is based on biased random walks to sample context information about each node in the network. Our random-walk based sampling can behave as Breath-First-Search (BFS) and Depth-First-Search (DFS) samplings with the goal to capture homophily and role equivalence between the nodes in the network. We have performed a detailed experimental evaluation comparing the performance of the proposed algorithm against various baseline methods, on several datasets and learning tasks. The experiment results show that the proposed method outperforms the baseline ones in most of the tasks and datasets.

\end{abstract}


%
\IEEEpeerreviewmaketitle

\section{Introduction}

Networks (or graphs) are used to model data arising from a wide range of applications -- ranging from social network mining, to biology and neuroscience. Of particular interest here are applications that concern learning tasks over networks.
For example, in social networks, to answer whether two given users belong in the same social community or not, examining their direct relationship does not suffice -- the users can have many further characteristics in common, such as friendships and interests. Similarly, in friendship recommendations, in order to determine whether two unlinked users are similar, we need to obtain an informative representation of the users and their proximity -- that potentially is not fully captured by handcrafted features extracted from the graph.
Towards this direction, representation (or feature) learning algorithms are useful tools that can help us addressing the above tasks. Given a network, those algorithms embed it into a new space (usually a compact vector space), in such a way that both the original network structure and other ``implicit'' features of the network are captured.

\par In this paper, we are interested in unsupervised (non-task-specific) feature learning methods; once the vector representations are obtained, they can be used to deal with various data mining tasks on the network, including node classification, community detection and link prediction by existing machine learning tools. Nevertheless, it is quite challenging to design an ``ideal" algorithm for network embeddings. The main reason is that, in order to learn task-independent embeddings, we need to preserve as many important properties of the network as possible in the embedding feature space. But such properties might be not apparent and even after they are discovered, preserving some of them is intractable.

This can explain the fact that most of the traditional methods only aim at retaining the first and the second-order proximity between nodes \cite{isomap00}, \cite{lle00}, \cite{laplacian02}. In addition to that, network representation learning methods are expected to be efficient so that they can scale well on large networks, as well as are able to handle different types of rich network structures, including (un)weighted, (un)directed and labeled networks.

\par In this paper, we propose BiasedWalk, an unsupervised and Skip-gram-based \cite{skipgram13} network representation learning algorithm which can preserve higher-order proximity information, as well as is able to capture both the homophily and role equivalence relationships between nodes. BiasedWalk relies on a novel node sampling procedure based on biased random walks, that can behave as actual depth-first-search (DFS) and breath-first-search (BFS) explorations -- thus, forcing the sampling scheme to capture both role equivalence and homophily relations between nodes. Furthermore, BiasedWalk is scalable on large scale graphs, and is able to handle different types of network structures, including (un)weighted and (un)directed ones. Our extensive experimental evaluation indicates that BiasedWalk outperforms the state-of-the-art methods on various network datasets and learning tasks.



\par Code and datasets for the paper are available at: 
\href{https://goo.gl/Easwk4}{{\color{blue}{\textbf{https://goo.gl/Easwk4}}}}.


\section{Related Work} \label{sec:related}
Network embedding methods have been well studied since the 2000s. The early works on the topic consider the embedding task as a dimensionality reduction one, and are often based on factorizing some matrix associated to pairwise distances between all nodes (or data points) \cite{isomap00,lle00,laplacian02}. 
Those methods rely on the eigen-decomposition of such a matrix are often sensitive to noise (i.e.\ missing or incorrect edges).
Some recently proposed matrix factorization-based methods focus on the formal problem of representation learning on networks, and are able to overcome limitations of the traditional manifold learning approaches. For example, while traditional ones only focus on preserving the first-order proximity, GraRep \cite{grarep15} aims to preserve a higher-order proximity by using many matrices, each of them is for $k$-step transition probabilities between nodes. The HOPE algorithm \cite{hope16} defines some similarity measures between nodes which are helpful for preserving higher-order proximities as well and formulates those measures as a product of sparse matrices to efficiently find the latent representations.


\par Recently, there is a category of network embedding methods that rely on representation learning techniques in natural language processing (NLP) \cite{deepwalk14}, \cite{tadw2015}, \cite{node2vec16}, \cite{revisiting2016}, \cite{mmdeepwalk2016}, \cite{dynamic2018}. The two best-known methods here are DeepWalk \cite{deepwalk14} and node2vec \cite{node2vec16}. Which both are based on the Skip-gram model \cite{skipgram13}. The model has been proposed to learn vector representations for words in a corpus by maximizing the conditional probabilities of predicting contexts given the vector representations of those words.
DeepWalk \cite{deepwalk14} is the first method to leverage the Skip-gram model for learning representations on networks, by extracting truncated random walks in the network and considering them as sentences. The sentences are then fed into the Skip-gram model to learn node representations. The node2vec algorithm \cite{node2vec16} can essentially be considered as an extension of Deepwalk. In particular, the difference between those two methods concerns the node sampling approach: instead of sampling nodes by uniform random walks, node2vec uses biased (and second-order) random walks to better capture the network structure.
Because the Skip-gram based methods use random walks to approximate high order proximities, their main advantages include both scalability and the ability to learn latent features that only appear when we observe a high order proximity (\textit{e.g.,} the community relation between nodes).

\par As we will present later on, the proposed method which is called BiasedWalk, is also based on the Skip-gram model. The core of BiasedWalk is a node sampling procedure which is able to generate biased (and first-order) random walks that behave as actual depth-first-search (DFS) and breath-first-search (BFS) explorations -- thus, helping towards capturing important structural properties, such as community information and structural roles of nodes. It should be noted that node2vec \cite{node2vec16} also aims to sample nodes based on DFS and BFS random walks, but it cannot control the distances between sampling nodes and the source node. Controlling the distances is crucial as in many specific tasks, such as the one of link prediction, nodes close to the source should have higher probability to form a link from it. Similarly, in the community detection task, given a fixed sampling budget we prefer to visit as many community-wide nodes as possible.

Another shortcoming of node2vec is that its second-order random walks require storing the interconnections between the neighbors of every node. More specifically, node2vec's space complexity is $\mathcal{O}(\tilde{D} \cdot |E|)$, where $\tilde{D}$ is the average degree of the input network $G = (V, E)$ \cite{node2vec16}. This can be an obstacle towards embedding dense networks, where the node degrees can be very high. Moreover, we prefer random-walk based sampling rather than pursuing pure DFS and BFS sampling because of the Markovian property of random walks. Given a random walk of length $L$, we can immediately obtain $L-k$ context sets for nodes in the random walk by sliding a window of size $k$ along the walk.


\par The LINE algorithm \cite{line15}, though not belonging to any of the previous categories, is also one of the well-known network embedding methods. LINE derives two optimization functions for preserving the first and the second order proximity, and performs the optimizations by stochastic gradient descent with edge sampling. Nevertheless, in general, the performance tends to be inferior compared to Skip-gram based methods. The interesting reader may refer to the following review articles on representation learning on networks \cite{DBLP:journals/corr/abs-1709-05584}, \cite{gem2018}. Lastly, we should mention here the recent progress on deep learning techniques that have also been adopted to learn network embeddings \cite{sdne16, dngr16, deeplearning4, ANRL2018, deeplearning5, deeplearning6}.

\section{Problem Formulation} \label{sec:formulation}
A network can be represented by a graph $G = (V,E)$, where $V$ is the set of nodes and $E$ is the set of links in the network. Depending on the nature of the relationships, $G$ can be (un)weighted and (un)directed. Table \ref{tab:TableOfNotation} includes notation used throughout the paper.

\begin{table}[t]\caption{Table of notation.}
\begin{center}
\begin{tabular}{l p{6cm} }
\toprule

$G = (V, E)$ & An input network with set $V$ of nodes and set $E$ of links \\
$\Phi(u)$ & Vector representation of $u$ \\
$C(u)$ & Set of neighborhood nodes of $u$ \\
$\Phi^\prime(w)$ & Vector representation of $w$ as it is in the role of a context node \\
$p_v$ & Probability of $v$ being the next node of the (uncompleted) random walk \\
$\tilde{D}$ & Average degree of the input network \\
$\alpha$ & Parameter for controlling the distances from the source node to sampled nodes \\
$L$ & Maximum length of random walks \\
$\tau_v$ & Proximity score of node $v$ \\
$\gamma$ & The number of sampled random walks per node \\
$N(u)$ & Set of neighbors of node $u$ \\
$c$ & Window size of training context of the Skip-gram \\
$\Psi(u, v)$ & Hadamard operator on vector representations of $u$ and $v$ \\

\bottomrule
\end{tabular}
\end{center}
\label{tab:TableOfNotation}
\end{table}


An \textit{embedding} of $G = (V, E)$ is a mapping from the node set to a low-dimensional space, $\Phi: V \rightarrow \mathbb{R}^d$, where $d \ll |V|$. 
Network embedding methods are expected to be able to preserve the structure of the original network, as well as to learn latent features of nodes. In this work, inspired by the Skip-gram model, we aim to find embeddings towards predicting neighborhoods (or ``context'') of nodes: 

\begin{equation}\label{eq1}
	\argmax\limits_{\Phi} {\sum\limits_{u \in V} \log p(C(u) \mid \Phi(u))}
\end{equation}

\noindent where $C(u)$ denotes the set of neighborhood nodes of $u$. (Note that, $C(u)$ is not necessarily a set of immediate neighbors of $u$, but it can be any set of sampled nodes for $u$. The insight is that nodes sharing the same context nodes are intended to be close enough in the graph or to be similar). For example, $C(u)$ could be a set of nodes within $k$-hop distance from $u$, or nodes that co-appear with $u$ in a short random walk over the graph. Some related studies have shown that the consequent embedding result is strongly affected by the adopted strategy for sampling the context nodes \cite{deepwalk14}, \cite{node2vec16}. For the purpose of making the problem tractable, we also assume that predicting nodes in a context set is independent of each other, thus Eq. \eqref{eq1} can be expressed as:

\begin{equation}\label{eq2}
	\argmax\limits_{\Phi} {\sum\limits_{u \in V} \sum\limits_{v \in C(u)} \log p(v \mid \Phi(u))}
\end{equation}

\noindent  Lastly, we use the softmax function to define the probability of predicting a context node $v$ given vector representation $\Phi(u)$ of $u$, as $p(v \mid \Phi(u)) = \frac{ \exp({\Phi^\prime(v)}^\intercal \Phi(u)) }{ \sum\limits_{w \in V} \exp({\Phi^\prime(w)}^\intercal \Phi(u)) }$, where $\Phi^\prime(w)$ is the context vector representation of $w$. Note that, there are two representation vectors for each $w \in V$: $\Phi(w)$ as $w$ is in the role of a target node, and $\Phi^\prime(w)$ with $w$ considered as a context node \cite{skipgram13}.

\section{Proposed Methodology} \label{sec:methodology}
\begin{figure}[t]
    \centering
    \begin{tabular}{ccc}
    \includegraphics[width=.2\textwidth]{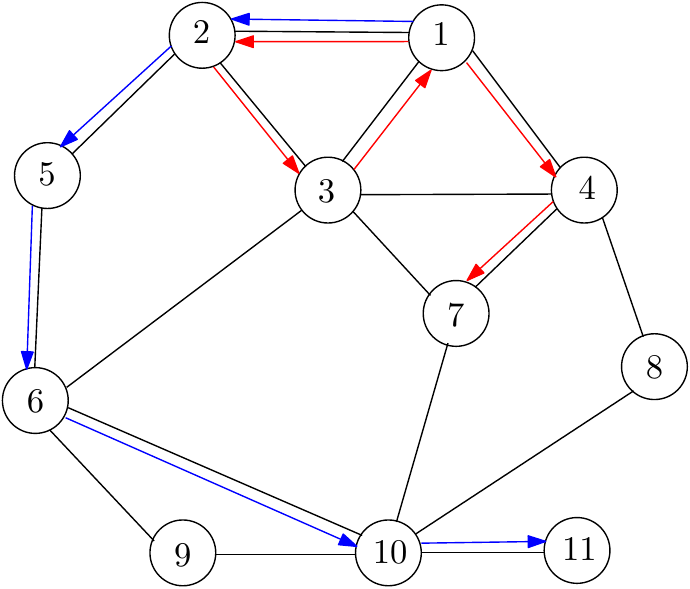} &~& 				\includegraphics[width=.2\textwidth]{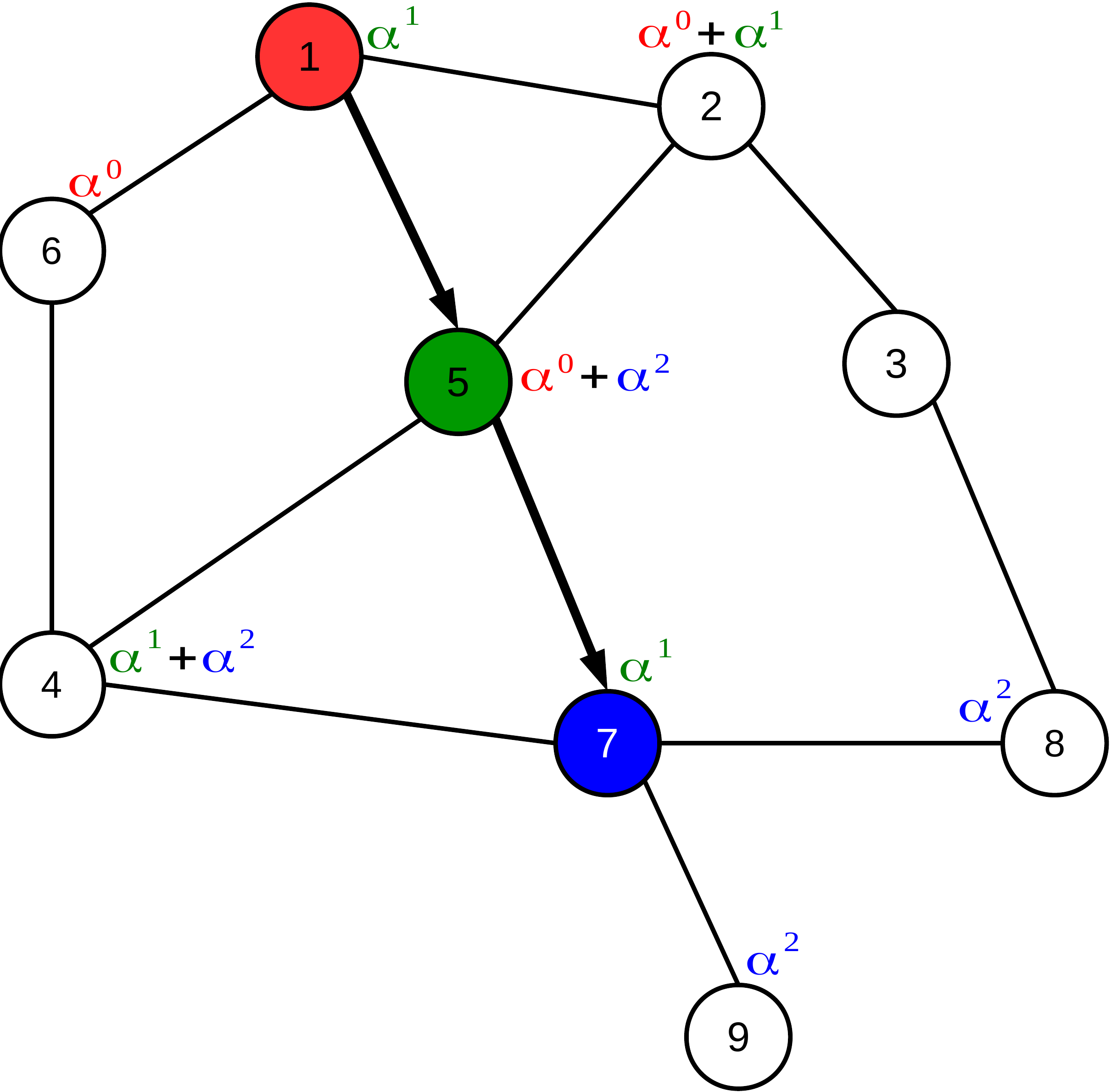} \\
    (a) &~& (b)
    \end{tabular}
\caption{(a) An example of BFS- and DFS-behavior random walks. \textmd{The two walks in this example both start from node 1. The BFS walk 1-2-3-1-4-7 (red) consists of many nodes in the local neighborhood of the source node, while nodes in the DFS walk 1-2-5-6-10-11 (blue) belong to a much wider portion of the network.} (b) An illustration of BiasedWalk's random-walk based sampling strategy. \textmd{The (uncompleted) walk sequence here is 1-5-7. Every node that has at least one neighbor visited, owns a proximity score that is shown near that node. For each term in a score sum, the node ``giving'' the amount of score is indicated by the same color. The next node for the walk will be one of the neighbors of node 7. In case of DFS-style sampling, nodes 8 and 9 should have higher probability of being selected as the next node (as their proximity scores are lower) compared to nodes 4 and 5.}}
    \label{fig:walks}
\end{figure}

In this section, we introduce BiasedWalk, a method for learning representation for nodes in a network based on the Skip-gram model. To achieve the goal for both capturing the network structure effectively and leveraging the efficient Skip-gram model for learning vector representation, we first propose a novel sampling procedure based on biased random-walks that are able to behave as actual depth-first-search (DFS) and breath-first-search (BFS) sampling schemes. Figure \ref{fig:walks} (a) shows an example of our biased random walks. On a limited budget of samples, random walks staying in the local neighborhood of the source node can be an alternative for BFS sampling \cite{node2vec16}, and such BFS random walks help in discovering the role of source nodes. For example, hub, leaf and bridge nodes can be determined by their neighborhood nodes only. On the other hand, random walks which move further away from the source node are equivalent to DFS sampling since such DFS walks could discover nodes in the community-wide area of the source, and this is helpful to understand the homophily effect between the nodes.

\par Given a random walk $u_1, u_2,..., u_L$, the contexts of $u_i$ are nodes in a window of size $c$ centered at that node, i.e., $C(u_i) = \{u_j \mid -c \leq j-i \leq c, j \neq i\}$. Then, we consider each of generated random walks as a sentence in a corpus to learn vector representations for words (\textit{e.g.,} nodes in the random walks) which maximize Eq. \eqref{eq2}. It should be noted here that the efficiency of each Skip-gram based method depends on its own sampling strategy. Our method, instead of uniform random walks (as performed by Deepwalk), uses biased ones for better capturing the network structure. Furthermore, it does not adopt the second-order random walks like node2vec since they are not able to control the distances between the source node and sampled nodes.

To be able to simulate both DFS and BFS explorations, BiasedWalk uses additional information $\tau_v$, called \textit{proximity score,} for each candidate node $v$ of the ongoing random walk, in order to estimate how far (not by the exact number of hops) a candidate is from the source node. More specifically, the nodes whose all neighbor nodes have never been visited by the random walk, should have a proximity score of zero. After the $i$-th node in the walk is discovered, the proximity score of every node adjacent to that node will be increased by $\alpha^{i-1}$, where $0 < \alpha \leq 1$ is a parameter \footnote{The parameter is to control the distances from the source to sampled nodes. Note that, in case $G$ is directed, we increase the proximity scores for all of the in- and out-neighbors of that node.}. Then, the probability distribution of selecting the next node for the current walk is calculated based on the proximity scores of the neighbor nodes of the most recently visited node, and on which type of samplings (DFS or BFS) we desire. In the case of BFS, the probability of a node being the next node should be proportional to its proximity score, i.e., $p_v = \frac{\tau_v}{\sum\limits_{w \in N(u)}{\tau_w}}$. In the case of DFS, the probability should be inversely proportional to that score, i.e., $p_v = \frac{\sfrac{1}{\tau_v}} {\sum\limits_{w \in N(u)} {\sfrac{1}{\tau_w}}}$, where $u$ is the most recently visited node and $N(u)$ defines the set of neighbor nodes of $u$. An illustration of our random-walk based sampling procedure is given in Figure \ref{fig:walks} (b), and its main steps are presented in Algorithm \ref{alg1}.


\par The reason for using an exponential function of the current walk length for increasing the proximity scores, is that it helps to clearly distinguish candidates that belong to the local neighborhood of the source node and others far away from the source. Since $0 < \alpha < 1$, those on the local neighborhood should have much higher scores than the ones outside. In addition, our exponential function guarantees that candidates of the same level of distance from the source, should have comparable proximity scores. Thus, such proximity scores are a good estimate of distances from the source node to candidates, and therefore help selecting the next node for our desired (DFS or BFS) random walks. 

\par Algorithm \ref{alg2} depicts the complete procedure of the proposed  BiasedWalk feature learning method. Given a desired sampling type (DFS or BFS), the procedure first generates a set of random walks from every node in the network based on Algorithm \ref{alg1}. In order to adopt the Skip-gram model for learning representations for the nodes, it considers each of these walks (a sequence of nodes) as a sentence (a sequence of words) in a corpus. Finally, the Skip-gram model is used to learn vector representations for words in the ``network'' corpus, which aims to maximize the probability of predicting context nodes as mentioned in Eq. \eqref{alg2}.

\begin{algorithm}[t]
\KwIn{A network $G = (V, E)$, source node $s, sampling\_type, \alpha, L$}
\KwOut{A random walk of maximum length $L$ starting from node $s$}
    
$walk \leftarrow [s]$		\\
$l \leftarrow 1$		{\textbackslash* The current walk length *\textbackslash}		\\
\While {$(l < L \And N(walk[l]) \neq \emptyset)$}
{
        $\tau \leftarrow \{ \}$		\textbackslash* A map from node IDs to their proximity scores to keep track nodes which have at least one neighbor visited by the current walk *\textbackslash		\\
        $u \leftarrow walk[l]$		\\
        
        \ForEach{$v \in N(u)$}
        {
                \If ({\textbackslash* In case none of the neighbors of $v$ has been visited *\textbackslash}){$v \not\in \tau.keys()$}
                {
                        $\tau[v] \leftarrow \alpha^{l-1}$
                }
                \Else
                {
                        $\tau[v] \leftarrow \tau[v] + \alpha^{l-1}$
                }
        }
        
        \ForEach{$v \in N(u)$}
        {
                \If {$sampling\_type = BFS$}
                {
                        $p_v \leftarrow \frac{\tau[v]}{\sum\limits_{w \in N(u)}{\tau[w]}}$
                }
                \ElseIf {$sampling\_type = DFS$}
                {
                        $p_v \leftarrow \frac{\sfrac{1}{\tau[v]}} {\sum\limits_{w \in N(u)} {\sfrac{1}{\tau[w]}}}$
                }			
        }
        
        $z \leftarrow $ randomly select a node in $\{v | v \in N(u)\}$ with the probabilities $p_v$ calculated		\\
        Append node $z$ to $walk$		\\
        $l \leftarrow l + 1$		\\
}

\Return $walk$
\caption{Global Random Walk}\label{alg1}
\end{algorithm}

\begin{algorithm}[h]
\KwIn{A network $G = (V, E), sampling\_type, \alpha$, number of walks per node $\gamma$, maximum walk length $L$}
\KwOut{Vector representations of nodes in $G$}

$walk\_set \leftarrow [\ ]$		\\
\For ({\textbackslash* generate $\gamma$ random walks from each node *\textbackslash}){$loop = 1$ \bf{to} $\gamma$}
{
	\ForEach{$s \in V$}
	{
		$walk \leftarrow Global\_Random\_Walk(G, s, sampling\_type, \alpha, L)$		{\textbackslash* Alg. \ref{alg1} *\textbackslash}		\\
		Append $walk$ to $walk\_set$		\\
	}
}

Construct a corpus $T$ consisting of $\gamma.|V|$ sentences where each sentence corresponds to a walk in $walk\_set$. The vocabulary $W$ of this corpus contains $|V|$ words, each corresponding to a node in $G$		\\

Use the Skip-gram to learn representations of words in corpus $T$ with vocabulary $W$		\\

\Return Corresponding vector representations for nodes in $G$
\caption{BiasedWalk}\label{alg2}
\end{algorithm}

\paragraph{\textbf{Space and time-complexity}} Let $\tilde{D}$ be the average degree of the input network, and assume that every node in the network has a degree bounded by $\mathcal{O}(\tilde{D})$. For each node visited in a random walk of maximum length $L$, Algorithm \ref{alg1} needs to consider all neighbors of that node to update (or initialize) their proximity scores, and then calculate the transition probabilities. The algorithm uses a map $\tau$ to store proximity scores of such neighbor nodes, so the number of keys (node IDs) in the map is $\mathcal{O}(\tilde{D} \cdot L)$. Thus, the time for both accessing and updating a proximity score is in $\mathcal{O}(\log{\tilde{D}} + \log{L})$, as the map can be implemented by a balanced binary search tree. The algorithm needs to select at most $L$ nodes whose degree is $\mathcal{O}(\tilde{D})$ to construct a walk. Therefore, the number of updating and accessing operations is $\mathcal{O}(L \cdot \tilde{D})$, and thus the time complexity of Algorithm \ref{alg1} is $\mathcal{O}(L \cdot \tilde{D}(\log{\tilde{D}} + \log{L}))$. With respect to memory requirement, BiasedWalk requires only $\mathcal{O}(|E|)$ space complexity since it adopts the first-order random walks.

\section{Experimental Evaluation} \label{sec:experimental}
For the experimental evaluation, we use vector representations of nodes learned by BiasedWalk and compare with four baseline methods, including DeepWalk \cite{deepwalk14}, node2vec \cite{node2vec16}, LINE \cite{line15} and HOPE \cite{hope16} in the tasks of multilabel node classification and link prediction.
For each type of our random walks (DFS and BFS), the value $\alpha$ of BiasedWalk varies in the range of $\{0.125, 0.25, 0.5, 1.0\}$. Then, we use 10-fold cross-validation on labeled data to choose the best parameters (including the walk type and value $\alpha$) for each graph dataset.

\subsection{Network datasets}
Table \ref{table:1} provides a summary of network datasets used in our experiments. More specifically, network datasets for the multilabel classification task include the following:
\begin{itemize}
	\item \textsc{BlogCatalog} \cite{blogcatalog2009}: A network of social relationships between the bloggers listed on the BlogCatalog website. There are $10,312$ bloggers, $333,983$ friendship pairs between them, and the labels of each node is a subset of 39 different labels that represent blogger interests (\textit{e.g.,} political, educational).
	
	\item Protein-Protein Interaction (PPI) \cite{ppi08}: The network contains $3,890$ nodes, $38,739$ unweighted edges and $50$ different labels. Each of the labels corresponds to a biological function of the proteins.
	
	\item \textsc{IMDb} \cite{WS_KDD13}: A network of $11,746$ movies and TV shows, extracted from the Internet Movie Database (IMDb). A pair of nodes in the network is connected by an edge if the corresponding movies (or TV shows) are directed by the same director. Each node can be labeled by a subset of $27$ different movie genres in the database, such as drama, comedy, documentary and action. 
	
\end{itemize}

For the link prediction task, we use the following network datasets (we focus on the largest (strongly) connected components instead of the whole networks): 

\begin{itemize}
	\item \textsc{AstroPhy} collaboration \cite{snap14}: The network represents co-author relationships between 17,903 scientists in AstroPhysics. 
	
	
	\item \textsc{Election-Blogs} \cite{AG05}: This is a directed network of front-page hyperlinks between blogs in the context of the 2004 US election. In the network, each node represents a blog and each edge represents a hyperlink between two blogs. 
	
	\item Protein-Protein Interaction (PPI): The same dataset used in the node-classification task.
    
    \item \textsc{Epinions}\cite{epinions2003}: The network represents who-trust-whom relationships between users of the \textit{epinions.com} product review website.
\end{itemize}

\begin{table*}[t]
\centering
\caption{Network datasets. \label{table:1}}
\begin{tabular}{l rc rc cc c} 
	\toprule
 	Network & $|V|$ &~~~~& $|E|$ &~~~~& Type &~~~~& {\#}Labels \\
 	\midrule
 	\textsc{BlogCatalog} & 10,312 &~~~~& 333,983 &~~~~& Undirected &~~~~& 39 \\ 
 	\textsc{PPI} & 3,890 &~~~~& 38,739 &~~~~& Undirected &~~~~& 50 \\
 	\textsc{IMDb} & 11,746 &~~~~& 323,892 &~~~~& Undirected &~~~~& 27 \\
	\midrule
	\textsc{AstroPhy} & 17,903 &~~~~& 197,031 &~~~~& Undirected &~~~~&  \\
	\textsc{Election-Blogs} & 1,222 &~~~~& 19,021 &~~~~& Directed &~~~~&  \\
    \textsc{PPI} (for link prediction) & 3,852 &~~~~& 21,121 &~~~~& Undirected &~~~~&  \\
    \textsc{Epinions} & 75,877 &~~~~& 508,836 &~~~~& Directed &~~~~&  \\
	\bottomrule
\end{tabular}
\end{table*}

\subsection{Baseline methods}
We use DeepWalk \cite{deepwalk14}, node2vec \cite{node2vec16}, LINE \cite{line15} and HOPE \cite{hope16} as baselines for BiasedWalk. The number of dimensions of output vector representations $d$ is set to 128 for all the methods. 
To get the best embeddings for LINE, the final representation for each node is created by concatenating the first-order and the second-order representations each of 64 dimensions \cite{line15}.
The Katz index with decay parameter $\beta = 0.1$ is selected for HOPE's high-order proximity measurement, since this setting gave the best performance in the original article \cite{hope16}.
Similar to BiasedWalk, DeepWalk and node2vec belong to the same category of Skip-gram based methods.
The parameters for both the node sampling and optimization steps of the three methods are set exactly the same: number of walks per node $\gamma = 10$; maximum walk length $L = 80$; Skip-gram's window size of training context $c = 10$. Since node2vec requires the in-out and return hyperparameters $p, q$ for its second-order random walks, we have performed a grid search over $p, q \in \{0.25, 0.5, 1, 2, 4\}$ and 10-fold cross-validation on labeled data to select the best embedding -- as suggested by the experiment in \cite{node2vec16}. For a fair comparison, the total number of training samples for LINE is set equally to the number of nodes sampled by the three Skip-gram based methods ($\#Sample = |V| \cdot \gamma \cdot L$).

\subsection{Experiments on multilabel classification}
Multilabel node classification is a challenging task, especially for networks with a large number of possible labels. To perform this task, we have used the learned vector representations of nodes and an one-vs-rest logistic regression classifier using the \textsc{LibLinear} library with $L2$ regularization \cite{liblinear08}. The Micro-F1 and Macro-F1 scores in the 50\%-50\% train-test split of labeled data are reported in Table \ref{table:2}. The best scores gained for each network across the methods are indicated in bold, and the best parameter settings including the walk type and the corresponding value $\alpha$ for BiasedWalk have been shown in the last row. In our experiment (including the link prediction task), each Micro-F1 or Macro-F1 score reported was calculated as the average score on $10$ random instances of a given train-test split ratio. 

\par
\begin{figure*}[t]
\centering
\includegraphics[width=.98\textwidth]{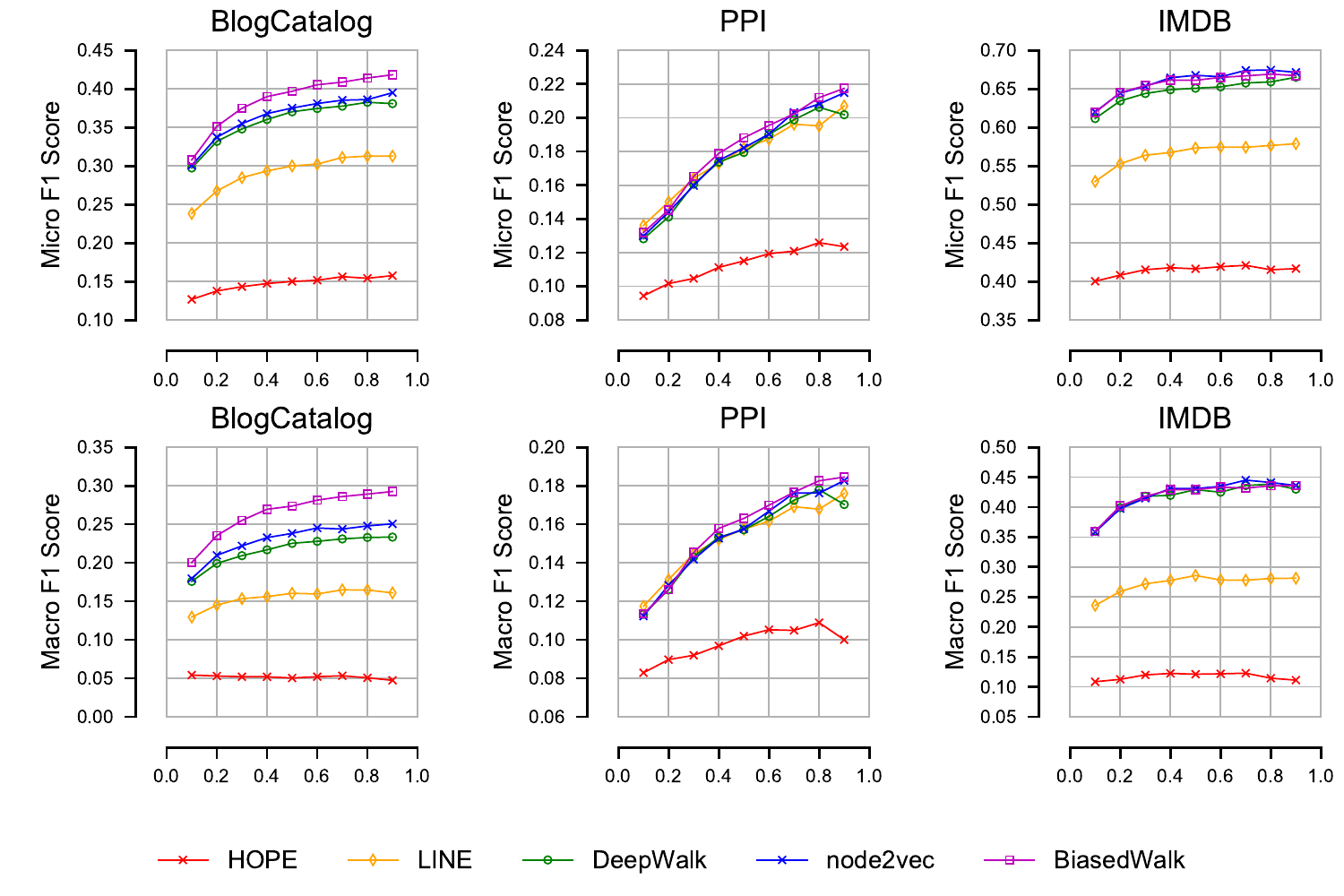}
\caption{Performance evaluation on various datasets for the node classification task.}
\label{fig:multi_label}
\end{figure*}

 The experimental results show that the Skip-gram based methods outperform the rest ones in the multilabel classification task. The main reason is that, the rest methods mainly aim to capture low order proximities for networks. But this is not enough for node classification as nodes which are not in the same neighborhood can be classified by the same labels. More precisely, BiasedWalk gives the best results for \textsc{BlogCatalog} and PPI networks; in \textsc{BlogCatalog}, it improves Macro-F1 score by 15\% and Micro-F1 score by 6\%. In \textsc{IMDb} network, node2vec and BiasedWalk are comparable, with the former being slightly better. Figure \ref{fig:multi_label} depicts the performance of all the methods in different percentages of training data. Given a percentage of training data, in most cases, BiasedWalk has better performance than the rest baselines. An interesting observation from the experiment is that, preserving homophily between nodes is important for the node classification task. This explains the reason why the best performance results obtained by BiasedWalk come from its DFS random-walk based sampling scheme.

\begin{table*}[t]
\centering
\caption{Macro-F1 and Micro-F1 scores (\%) for 50\%-50\% train-test split in the multi-label classification task. \label{table:2}}
\begin{tabular}{cccccccccc} 
	\toprule
 	\multirow{3}{*}{Algorithm} && \multicolumn{6}{c}{Networks} \\
 	& \multicolumn{3}{c}{\textsc{BlogCatalog}} && \multicolumn{2}{c}{\textsc{PPI}} && \multicolumn{2}{c}{\textsc{IMDb}} \\
 	 \cmidrule{3-4}  \cmidrule{6-7}  \cmidrule{9-10}
 	&~~& Macro & Micro &~~& Macro & Micro &~~& Macro & Micro \\
 	HOPE &~~& 5.04 & 15.00 &~~& 10.19 & 11.50 &~~& 12.11 & 41.64 \\
 	LINE &~~& 16.03 & 30.00 &~~& 15.75 & 18.19 &~~& 28.59 & 57.33 \\
    \midrule
 	DeepWalk &~~& 22.53 & 37.03 &~~& 15.71 & 17.94 &~~& 42.91 & 65.09 \\
 	node2vec &~~& 23.82 & 37.52 &~~& 15.76 & 18.22 &~~& \textbf{43.10} & \textbf{66.77} \\
 	BiasedWalk &~~& \textbf{27.36} & \textbf{39.69} &~~& \textbf{16.30} & \textbf{18.81} &~~& 42.91 & 66.12 \\
	(Best combination of: $walk\_type, \alpha$) &~~& \multicolumn{2}{c}{DFS, 1.0} &~~& \multicolumn{2}{c}{DFS, 0.5} &~~& \multicolumn{2}{c}{DFS, 0.25} \\
    \bottomrule
\end{tabular}
\end{table*}

\subsection{Experiments on link prediction}
To perform link prediction in a network dataset, we first randomly remove half of its edges, so that after the removal the network is not disconnected. The node representations are then learned from the remaining part of the network. To create the negative labels for the prediction task, we randomly select pairs of nodes that are not connected in the original network. The number of such pairs is equal to the number of removed edges. The ``negative'' pairs and the pairs from edges that have been removed, are used together to form the labeled data for this task.

Because the prediction task is for edges, we need some way to combine a pair of vector representations of nodes to form a single vector for the corresponding edge. Some operators have been recommended in \cite{node2vec16} for handling that. We have selected the Hadamard operator, as it gave the best overall results. 
Given vector representations $\Phi(u)$ and $\Phi(v)$ of two nodes $u$ and $v$, the Hadamard operator defines the vector representation of link $(u, v)$ as $\Psi(u, v)$, where $\Psi_i(u, v) = \Phi_i(u) \cdot \Phi_i(v), \forall i =1, \ldots, d$. Then, we use a Linear Support Vector Classifier with the $L2$ penalty to predict whether links exist or not.

The experimental results are reported in Table \ref{table:3}. Each score value shows the average over $10$ random instances of a 50\%-50\% train-test split of the labeled data. In this task, HOPE is still inferior to the other methods, except in the \textsc{Election-Blogs} network. Since HOPE is able to preserve the asymmetric transitivity property of networks, it can work well in directed networks, such as the \textsc{Election-Blogs}. LINE's performance is comparable to that of the Skip-gram based methods and even the best one in the PPI network. The reason is that LINE is proposed to preserve the first and the second-order proximity and that is really helpful for the link prediction task. It is totally possible to infer the existence of an edge if we know the role of its nodes and meanwhile, roles of nodes can be discovered by examining just their local neighbors. BiasedWalk, again, gains the best results for this task in most of the networks. Finally, the best results of BiasedWalk in this task are almost based on its BFS sampling. This supports the fact that, discovering role equivalence between nodes is crucial for the link prediction task.

\begin{table*}[t]
\centering
\caption{Macro-F1 and Micro-F1 scores (\%) of the 50\%-50\% train-test split in the link prediction task. HOPE could not handle the large \textsc{Epinions} graph because of its high time and space complexity.}
\label{table:3}
\begin{tabular}{c c ccccc c c c cc c c}
	\hline
 	\multirow{4}{*}{Algorithm} && \multicolumn{8}{c}{Network} \\
 	& \multicolumn{3}{c}{\textsc{AstroPhy}} && \multicolumn{2}{c}{\textsc{Election-Blogs}} && \multicolumn{2}{c}{\textsc{PPI}} && \multicolumn{2}{c}{\textsc{Epinions}} \\
 	 \cmidrule{3-4}  \cmidrule{6-7}  \cmidrule{9-10} \cmidrule{12-13}
 	&~~& Macro~~ & Micro &~~& Macro & Micro &~~& Macro & Micro &~~& Macro~~ & Micro \\
 	HOPE &~~& 59.97~~ & 65.61 &~~& 76.34 & 76.99 &~~& 53.15~~ & 61.45 &~~& *~~ & * \\
 	LINE &~~& 93.56~~ & 93.59 &~~& 74.79 & 74.80 &~~& \textbf{80.29}~~ & \textbf{80.34} &~~& 81.94 & 81.95 \\
    \midrule
 	DeepWalk &~~& 92.29~~ & 92.32 &~~& 79.06 & 79.12 &~~& 67.84~~ & 67.99 &~~& 89.14 & 89.18 \\
 	node2vec &~~& 93.08~~ & 93.10 &~~& 81.04 & 81.15 &~~& 71.24~~ & 71.29 &~~& 89.29 & 89.33 \\
 	BiasedWalk &~~& \textbf{95.08} ~~& \textbf{95.10} &~~& \textbf{81.54} & \textbf{81.65} &~~& 75.23~~ & 75.34 &~~& \textbf{90.08} & \textbf{90.11} \\
	 (The best $walk\_type, \alpha$) &~~& \multicolumn{2}{c}{BFS, 0.125} &~~& \multicolumn{2}{c}{BFS, 0.125} &~~& \multicolumn{2}{c}{BFS, 1.0} &~~& \multicolumn{2}{c}{BFS, 0.25} \\
    \hline
\end{tabular}
\end{table*}

\subsection{Parameter sensitivity and scalability}
\begin{figure*}[t]
\centering
\includegraphics[width=.73\textwidth]{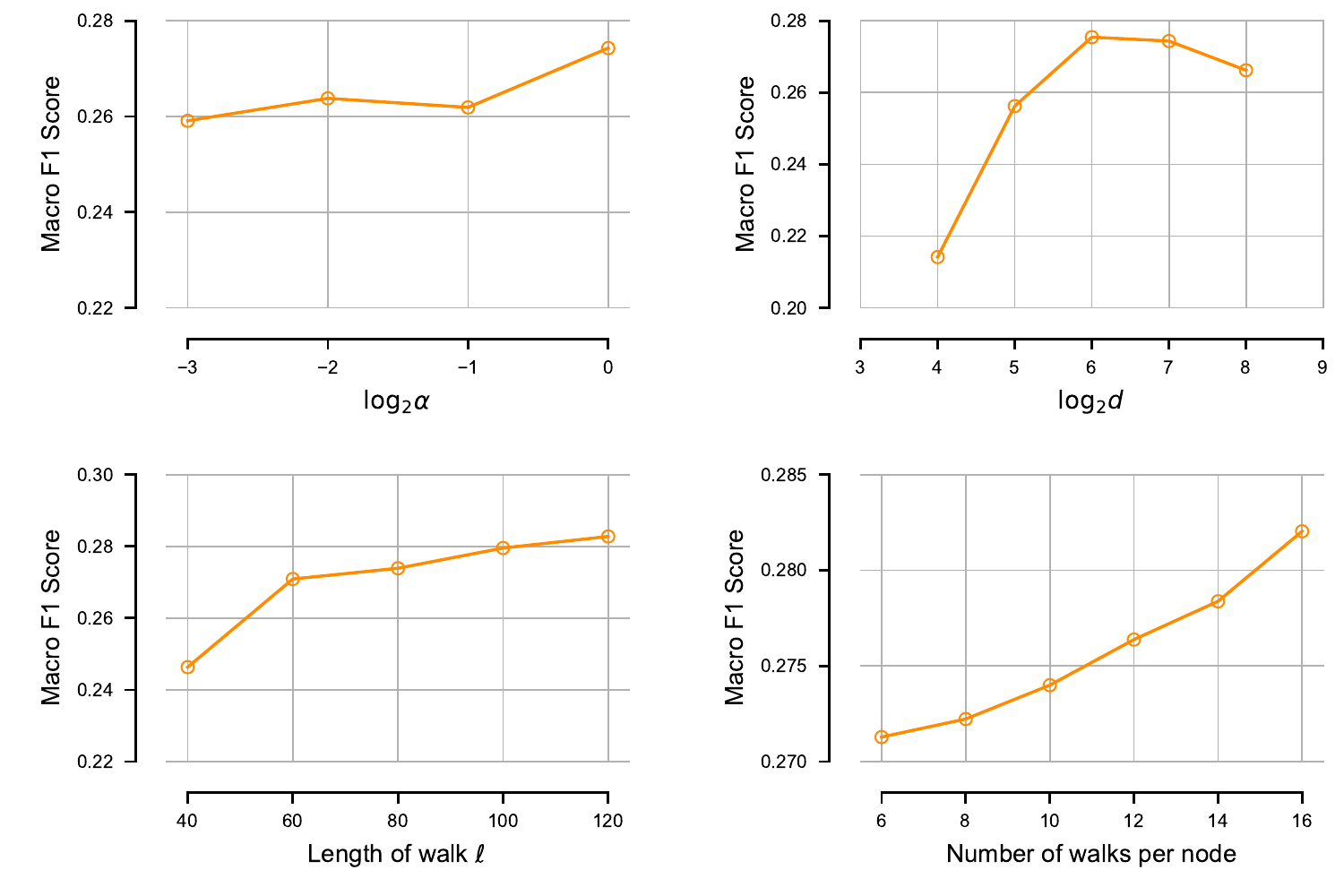}
\caption{Parameter sensitivity of BiasedWalk in \textsc{BlogCatalog}. \label{fig:parameter}}
\end{figure*}

We also evaluate how BiasedWalk's performance is changing under different parameter settings. Figure \ref{fig:parameter} shows Macro-F1 scores gained by BiasedWalk in the multilabel classification task in \textsc{BlogCatalog} (the Micro-F1 scores follow a similar trend then it is not necessary to show them here). Except the parameter is being considered, all other parameters in the experiment are set to their default value. Obviously, since the number of walks per node or the walk length is increased, there are more nodes sampled by the random walks and then BiasedWalk should get a better score. The effectiveness of BiasedWalk also depends on the dimension number of output vector representations $d$. Since the number of dimensions is too small, embeddings in the representation space may not be able to preserve the structure information of input networks, and as this parameter is set so high it could negatively affect consequent classification tasks. Finally, we can notice the dependence of Macro-F1 score and the tendency of parameter $\alpha$, this can support us in inferring the best value for the parameter on each network dataset.

\par
\begin{figure}[h]
  \begin{center}
  \includegraphics[width=0.4\textwidth]{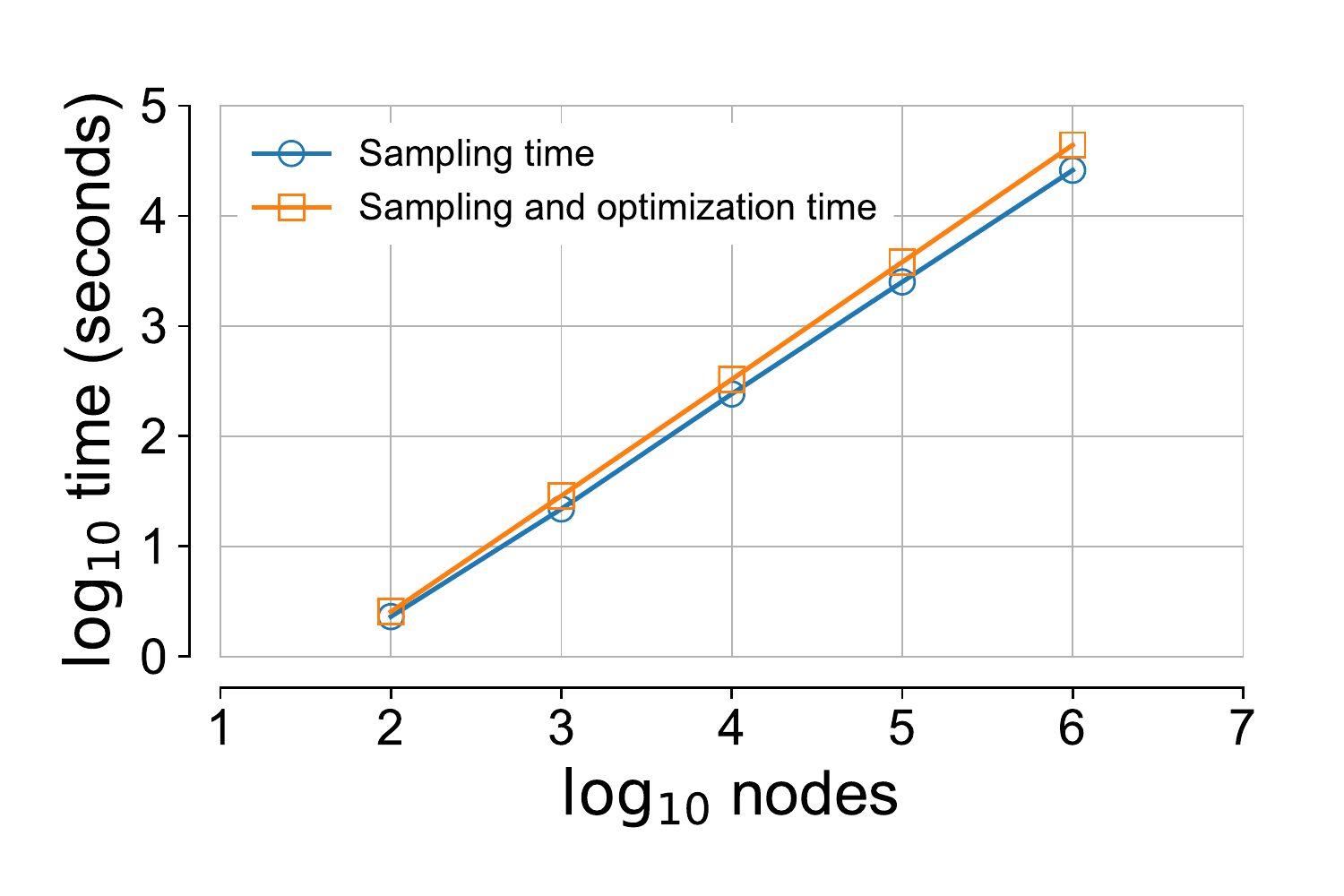}
  \end{center}
  \caption{Scalability of BiasedWalk on Erd\H{o}s-R\'{e}nyi  graphs. \label{fig:scalability}}
\end{figure}

 We have also examined the efficiency of the proposed BiasedWalk algorithm.  Figure \ref{fig:scalability} depicts the running time required for sampling (blue curve) and both sampling and optimization (orange curve) on the Erd\H{o}s-R\'{e}nyi graphs of various sizes ranging from $100$ to $1M$ nodes with the average degree of 10. As we can observe, BiasedWalk is able to learn embeddings for graphs of millions of nodes in dozens of hours and scales linearly with respect to the size of graphs. Nearly total learning time belongs to the step of sampling nodes that means the Skip-gram is very efficient at solving Eq. \eqref{eq2}.

\section{Conclusions and Future Work} \label{sec:conclusions}
In this work, we have proposed BiasedWalk, a Skip-gram based method for learning node representations on graphs. The core of BiasedWalk is a node sampling procedure using biased random-walks, that can behave as actual depth-first-search and breath-first-search explorations -- thus, forcing BiasedWalk to efficiently capture role equivalence and homophily between nodes. We have  compared BiasedWalk to several state-of-the-art baseline methods, demonstrating its good performance on the link prediction and multilabel node classification tasks. 
As future work, we plan to theoretically analyze the properties of the proposed biased random-walk scheme and to investigate how to adapt the scheme for networks with specific properties, such as signed networks and ego networks, in order to obtain better embedding results.



%

\balance
\bibliographystyle{splncs_srt}
\bibliography{ref}

\end{document}